\documentclass[letterpaper, 10 pt, conference]{ieeeconf}  
\usepackage{url}
\let\labelindent\relax
\usepackage{enumitem}
\usepackage{booktabs}
\usepackage{placeins}
\usepackage{multirow}
\usepackage{graphicx}
\usepackage{threeparttable}
\usepackage{amsmath,amssymb}
\usepackage{xcolor}

\IEEEoverridecommandlockouts                              

\title{\LARGE \bf
Shape-Aware Neural Time Fields for Rigid-Body Motion and Manipulation Planning
}

\author{Yanming Zhou$^{*}$, Yuchen Liu$^{*}$, and Ahmed H. Qureshi
\thanks{$^{*}$denotes equal contributions}%
\thanks{All authors are with the Department of Computer Science, Purdue University, West Lafayette, IN 47907, USA.
{\tt\small \{zhou1530, liu3853, qureshi7\}@purdue.edu}}%
}

\usepackage{etoolbox}
\makeatletter
\newcommand{\paperteaser}{%
\begin{minipage}{\textwidth}\def\@captype{figure}
\centering
\includegraphics[width=\textwidth]{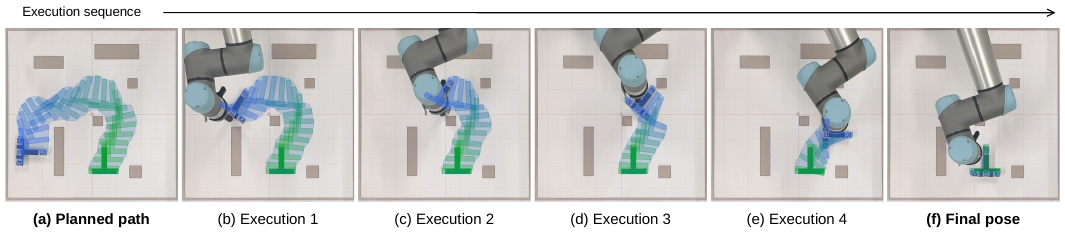}
\caption{Shape-NTFields enables closed-loop UR5e pushing with fast trajectory inference (0.048 s on average in the 2D benchmark). (a) Planned T-block path around fixed obstacle regions. (b) to (e) Intermediate execution with updated paths. (f) Final observed pose. The system measures the block pose and replans after each short push.}
\label{fig:real-pusht}
\end{minipage}\par\vspace{3mm}
}
\apptocmd{\@maketitle}{\paperteaser}{}{\PackageError{paper}{Cannot insert opening figure}{Check title definition.}}
\makeatother

\begin{document}
\makeatletter
\immediate\write\@auxout{\string\citation{BSTcontrolNoAuthorDash}}
\makeatother

\maketitle
\setcounter{figure}{1}
\thispagestyle{empty}
\pagestyle{empty}

\begin{abstract}
Rigid-body motion planning is fundamental to robotic transport and manipulation, where both object position and orientation determine whether a body can move through cluttered environments. This coupling makes planning challenging because translation and rotation can be constrained very differently by surrounding geometry. We propose Shape-Aware Neural Time Fields, a physics-informed neural motion planning framework that learns a reusable travel-time, or cost-to-go, representation for rigid-body motion while explicitly accounting for these distinct geometric restrictions. The learned field captures local motion difficulty in translation and rotation and is paired with a sampling-based planner that balances progress toward the goal with the cost of reaching nearby poses. We evaluate the framework across 2D and 3D rigid-body planning tasks and non-prehensile manipulation through object pushing in diverse environments. Results show substantial improvements in planning success and inference efficiency over physics-informed neural planning baselines, while remaining competitive with classical sampling-based planners at significantly lower planning times. Ablation studies further validate the contributions of both the learned representation and the proposed planning strategy. Finally, closed-loop pushing experiments under execution deviations and external disturbances demonstrate that fast replanning enables reactive and adaptive task execution. Video demo can be found here: https://youtu.be/5rUwvolPAgE. Code implementation can be found here: https://github.com/Ymz2006/Shape-NTFields
\end{abstract}

\section{Introduction}
\label{sec:introduction}

The piano movers' problem asks how to move a rigid body between two poses without colliding with obstacles~\cite{piano-movers}. In object transport and manipulation, both position and orientation determine whether a body can pass through openings and clutter. Rigid-body planning must therefore account for translation and rotation, which geometry can constrain differently.

Classical planners use search, sampling, or trajectory optimization~\cite{Hart1968Formal,LaValle1998Rapidly-exploring,Kuffner2000RRT-connect:,Ratliff2009CHOMP,Kalakrishnan2011STOMP:}. Many solve each start-goal query separately. PRM and LazyPRM reuse roadmaps but still require online connections and graph search~\cite{Kavraki1996PRM,Bohlin2000LazyPRM}, and the resulting paths may require further optimization~\cite{Schulman2014TrajOpt}. These recurring costs motivate reusable planning representations with efficient inference.

Learning-based planners shift part of this computation offline by reusing trained models to predict paths or planning costs for new start-goal queries~\cite{Qureshi2019Motion,Dalal2025NeuralMP}. Physics-informed approaches such as Neural Time Fields (NTFields) instead learn continuous goal-conditioned travel-time representations directly from geometry without requiring expert trajectories~\cite{ni2023ntfields,Ni2023Progressive}. Geometry sets a scalar propagation speed that measures local motion difficulty, rather than physical execution velocity: greater clearance gives higher speed, while proximity to obstacles lowers it. The learned field provides reusable goal-dependent costs for motion generation. Recent work extends this approach through metric learning and planning in large or unknown environments~\cite{ni2025physics,Liu2025Physics-informed,Liu2024Physics-Informed}.

For rigid bodies, however, a single geometric notion of local motion difficulty may be insufficient because translation and rotation are restricted differently by obstacles. An elongated object may translate along a narrow corridor yet collide with a wall when attempting to turn. Reaching the next opening may therefore require first moving into free space, reorienting, and then approaching the passage. Existing shape-aware and object-pose neural time fields account for body geometry and orientation~\cite{Shen2024PC-Planner:,Ren2025Physics-informed}, but their speed supervision does not separately specify translational and angular clearance. We therefore introduce separate translation and rotation representations and geometry-derived propagation targets, with both targets depending on the complete pose to retain their geometric coupling.

Reliable path extraction must also account for local motion difficulty. A candidate with low remaining travel time may still be difficult to reach through a constrained translation or rotation. We therefore score both its remaining time to the goal and the local cost of reaching it. Variations in field-derived speed across sampled motions then guide a steering correction, connecting geometric supervision to path generation.

We introduce Shape-Aware Neural Time Fields (\textbf{Shape-NTFields}), combining distinct translation and rotation supervision with sampling-based path extraction. For each fixed body and environment, the field serves new start-goal queries and closed-loop replanning from updated poses.

\noindent Our contributions are as follows:
\begin{itemize}[leftmargin=*]
    \item \textbf{Translation and rotation supervision.} Separate representations and geometric speed targets capture the two motion components within one pose-dependent field. Inverse-speed weighting emphasizes constrained poses, and Relative Normal Loss supervises speed decrease toward reduced clearance.
    \item \textbf{Planning with local and global travel-time costs.} Our sampling-based planner balances remaining travel time against local motion cost. Speed variation across candidates supplies a steering direction for the next pose update.
\end{itemize}
  
  We evaluate the framework on 2D and 3D rigid-body planning tasks spanning seven shapes and four environments per dimension, together with closed-loop non-prehensile pushing experiments. In 3D, our method achieves 92.9\% success with 0.055 s mean inference time, compared with 63.7\% success for TDM~\cite{ni2025physics}, while the 2D planner achieves 94.3\% success at 0.048 s mean inference time. Qualitative real-world UR5e demonstrations further show repeated replanning after execution deviations and external perturbations.

\section{Related Work}
\label{sec:related-work}

\subsection{Classical and Learning-based Motion Planning}
On configuration-space grids, search and Eikonal solvers face discretization costs that grow rapidly with dimension~\cite{Hart1968Formal,Sethian1996FastMarching}. Sampling-based planners require collision checks, and their paths may need smoothing~\cite{Kuffner2000RRT-connect:,Karaman2011Sampling}. Roadmaps reuse connections across queries but still require graph search~\cite{Kavraki1996PRM,Bohlin2000LazyPRM}. Trajectory optimization directly refines paths, but convergence and solution quality depend on initialization~\cite{Ratliff2009CHOMP,Kalakrishnan2011STOMP:,Schulman2014TrajOpt,Dong2016GPMP2}.

To reduce these online costs, learning-based methods reuse experience across queries by predicting sampling distributions~\cite{Ichter2018Sampling} or successive configurations~\cite{Qureshi2019Motion}. Motion Policy Networks and Neural MP learn reactive motion generation from expert plans, with Neural MP adding test-time optimization~\cite{Fishman2022Motion,Dalal2025NeuralMP}. These methods require expert motion data.

\subsection{Shape-aware Motion Planning}
Shape-aware planning must account for the entire body as its position and orientation change. Swept-volume SDF methods represent the region occupied along a trajectory and optimize continuous collision avoidance for arbitrary shapes~\cite{Wang2024Implicit,Li2025Efficient,Wang2025Any-shape}. These methods optimize each query's trajectory. Another route learns reusable geometry through collision classification~\cite{Das2017Fastron}, differentiable collision scoring~\cite{Zhi2022DiffCo}, or separation-distance estimation~\cite{Kew2020ClearanceNet}. Configuration-space distance fields provide distances and gradients to contact or collision boundaries~\cite{li2024configuration,Chen2026CSSDF-Net:}. These quantities support search and control but do not encode accumulated travel time around obstacles to a goal.

SAMP instead fuses robot and environment SDF features on spatial anchors with current and goal configurations to predict joint updates~\cite{chen2025sampspatialanchorbasedmotion}. It requires expert trajectories; we learn a goal-dependent field directly from geometry.

\subsection{Physics-informed Neural Motion Planning}
Physics-informed neural planning builds on differential-equation supervision~\cite{Raissi2019PINNs}, from NTFields' Eikonal formulation to progressive learning and temporal-difference metric learning~\cite{ni2023ntfields,Ni2023Progressive,ni2025physics}. Subsequent work scales neural time fields to large environments, incorporates sparse roadmap supervision, and supports online mapping and planning~\cite{Liu2025Physics-informed,Ni2026Weakly-supervised,Liu2024Physics-Informed,Chen2025Online}. Beyond these applications, Eikonal learning has been formulated on constraint manifolds and with Riemannian metrics~\cite{Ni2024ConstraintManifolds,Li2024Riemannian}, while Soner boundary conditions and sparse distance supervision improve geodesic learning~\cite{CabralMuchacho2025Soner}.

For rigid-body planning, PC-Planner uses shape-aware distance predictions for speed supervision and imposes monotonicity and optimality constraints~\cite{Shen2024PC-Planner:}. Neural time fields for prehensile manipulation encode object shape and pose, with obstacle distance and reachability determining speed targets and Dirichlet energy regularizing training~\cite{Ren2025Physics-informed}. Both account for rotation through pose-dependent geometry, but do not separately supervise translational and angular clearance. We provide a distinct target for each within one travel-time field.

\section{Background: Neural Time Fields}
\label{sec:ntfields-background}
In NTFields~\cite{ni2023ntfields}, $T(q,q_g)$ denotes travel time from configuration $q$ to goal $q_g$ in free space. For propagation speed $S(q)>0$, the standard isotropic Eikonal formulation is
\begin{equation}
S(q)\|\nabla_q T(q,q_g)\|_2=1,\qquad T(q_g,q_g)=0.
\label{eq:ntfields-background}
\end{equation}
Geometry-derived speeds decrease near obstacles; automatic differentiation of the neural prediction supplies gradients for Eikonal-residual training. These standard conditions motivate our blockwise supervision of an approximate planning cost. TDM~\cite{ni2025physics} introduces pairwise embeddings and temporal-difference supervision; we retain its embedding formulation and causality weighting.

\section{Methodology}
\label{sec:method}

\begin{figure*}[t]
\vspace*{2pt}
\centering
\includegraphics[width=\textwidth]{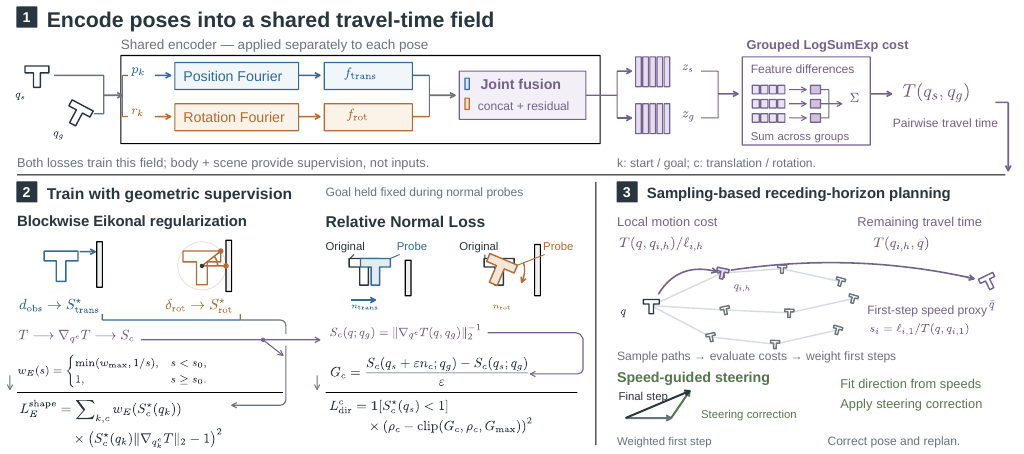}
\caption{Method overview. (1) Start and goal poses share an encoder with separate translation and rotation branches followed by fusion. Comparing their embeddings gives the approximate travel time $T(q_s,q_g)$. (2) Geometry supplies speed targets $S_c^\star$ and normals $n_c$ for translation and rotation ($c$). Gradients of $T$ give network-implied speeds $S_c$. Blockwise Eikonal regularization aligns them with the targets using inverse-speed weighting; Relative Normal Loss encourages speed decrease along $n_c$ with the goal fixed. Both losses train the same field. (3) The trained field scores candidates using remaining time $T(q_{i,h},\bar q)$ and local slowness $T(q,q_{i,h})/\ell_{i,h}$, the predicted time per displacement from the current pose. Scores weight first steps, while inverse slowness supplies speeds for fitting a steering direction. The planner corrects the weighted step and repeats, extending both endpoints until their paths connect.}
\label{fig:method-overview}
\end{figure*}

We present Shape-NTFields, a physics-informed framework that learns a reusable travel-time field over rigid-body poses and uses it for sampling-based path generation. The key idea is to separately model the
geometric restrictions on translation and rotation while retaining a shared pose-dependent travel-time
representation for planning. Figure~\ref{fig:method-overview} summarizes the framework.

\subsection{Problem Formulation}
Given a fixed rigid body $\mathcal B$ and a static obstacle region $\mathcal X_{\mathrm{obs}}$, we seek paths between collision-free configurations $q_s,q_g\in\mathcal Q_{\mathrm{free}}$, where $\mathcal Q_{\mathrm{free}}$ is the free subset of the configuration space $\mathcal Q$. We represent a spatial configuration as $q=(p,r)$, with position $p\in\mathbb R^3$ and normalized rotation vector $r=\omega/(2\pi)\in\mathbb R^3$, where $\omega$ encodes the rotation axis and angle. Positions are normalized by the environment scale. Planar tasks restrict motion to two positional coordinates and one rotation axis.

Our network receives only $(q_s,q_g)$ and predicts an approximate travel-time cost. Each model is trained for a fixed body and environment using geometric supervision, without ground-truth travel times, and reused across queries. The cost reflects motion difficulty through the speed targets: a short path through low-speed poses can be more costly than a longer path with greater clearance.

\subsection{Translation and Rotation Speed Targets and Normals}
Recall from the NTFields Eikonal relation in Eq.~(1) that the geometry-dependent propagation speed
determines the desired magnitude of the travel-time gradient. For rigid-body motion, however, a single
propagation-speed target does not explicitly distinguish whether motion is locally restricted by translation or
by rotation. We therefore derive separate translational and rotational speed targets from the geometry of the
body and obstacles.
Translational clearance $d_{\mathrm{obs}}$ is the minimum sampled obstacle-to-body distance. For
rotational clearance, we group body and obstacle samples by radial distance from the rotation center. We
approximate $\delta_{\mathrm{rot}}$ by the smallest angular gap among sample pairs within the same radial
group. This geometric surrogate measures limited room to turn; it is not an exact first-collision angle.

TDM uses a single configuration-level speed target derived from obstacle clearance and applies it to the full
configuration-space travel-time gradient. In contrast, we define separate translational and rotational targets,
each derived from its corresponding geometric clearance:
\begin{equation}
S_c^\star(q)=\max\!\left\{\psi\!\left(\operatorname{clip}\!\left(
\frac{d_c(q)}{d_{c,\max}},\frac{d_{c,\min}}{d_{c,\max}},1\right)\right),S_{\min}\right\},
\label{eq:target-speeds}
\end{equation}
where $c\in\{\mathrm{trans},\mathrm{rot}\}$ selects the motion component: $S_c^\star$ denotes either geometry-derived target $S_{\mathrm{trans}}^\star$ or $S_{\mathrm{rot}}^\star$, not their combination. Here $d_{\mathrm{trans}}=d_{\mathrm{obs}}$, $d_{\mathrm{rot}}=\delta_{\mathrm{rot}}$, and $d_{c,\min},d_{c,\max}$ are block-specific clearance thresholds. The function $\operatorname{clip}(x,l,u)$ truncates $x$ to $[l,u]$, $\psi(\eta)=\eta^2(2-\eta)^2$, and $S_{\min}>0$ is a speed floor. Both targets depend on the complete pose and decrease as clearance decreases. These scalar speeds do not prescribe a direction of travel.

We orient the two pose-space normals toward decreasing clearance, so positive probes encourage lower predicted speed. The translation normal $n_{\mathrm{trans}}$ points from the nearest body point toward the obstacle, with zero rotation. The rotation normal $n_{\mathrm{rot}}$ has zero translation and uses the normalized cross product of the body radial direction with the obstacle radial direction, in that order (body $\times$ obstacle). We apply it as a local perturbation in rotation-vector coordinates to encourage speed decrease toward reduced angular clearance. Each pair has collision-free endpoints, speed labels at both endpoints, and normals at the first; no connecting trajectory is required.

\subsection{Network Architecture and Training Objective}
The two speed targets describe different motion components, but a path can require both translation and rotation. We therefore distinguish their features and supervision while predicting one travel time for the complete pose pair.

\textbf{Separate Translation and Rotation Encoders.}
Our encoders separate translation and rotation features for their corresponding geometric speed supervision.
Whereas TDM represents the complete configuration using a single feature encoder $f_\theta(q)$, we first
encode translation and rotation through separate branches and then fuse their representations into a shared
pose embedding. TDM similarly represents travel time as a distance between learned start and goal
embeddings. Our formulation retains this pairwise metric structure while introducing the translation–rotation
decomposition required by our geometric supervision
\begin{equation}
\begin{aligned}
f_\theta(q)&=f_{\mathrm{fuse}}\left(
[f_{\mathrm{trans}}(\gamma_{\mathrm{trans}}(p));
f_{\mathrm{rot}}(\gamma_{\mathrm{rot}}(r))]\right),\\
T(q_s,q_g)&=D\left(f_\theta(q_s),f_\theta(q_g)\right).
\end{aligned}
\label{eq:shape-embedding}
\end{equation}
Here $\theta$ comprises the learned parameters, $\gamma_c$ is the Fourier feature map for block $c$, and $[\cdot\,;\,\cdot]$ denotes feature concatenation.. The two branch encoders have separate parameters and feed a residual fusion network $f_{\mathrm{fuse}}$. The complete encoder $f_\theta$ is shared across endpoints. Fusion retains the dependence of travel time on both motion components. Thus, the effect of an orientation change can depend on position.

\textbf{Grouped LogSumExp Cost.}
We retain TDM's grouped metric-learning structure, in which feature differences are aggregated within
groups and then summed across groups. TDM uses a within-group maximum, corresponding to an
$L_\infty$-type aggregation, followed by summation across groups. We replace this within-group maximum
with a smooth LogSumExp approximation while retaining the grouped structure. Each embedding $z_k=f_\theta(q_k)$, $k\in\{s,g\}$, contains $ab$ features, partitioned into $a$ groups with $b$ features per group. Let $\Delta_{ij}=|z_{s,ij}-z_{g,ij}|$. We use
\begin{equation}
D(z_s,z_g)=\frac{\kappa}{\beta}\sum_{i=1}^{a}
\log\!\left[\frac{1}{b}\sum_{j=1}^{b}
\exp(\beta\Delta_{ij})\right],
\label{eq:smooth-distance}
\end{equation}
where $i,j$ index groups and features, $\beta>0$ controls how closely LogSumExp approximates the maximum, and $\kappa>0$ scales the summed output. Numerical smoothing of the absolute difference is omitted from the notation. The resulting cost is symmetric and nonnegative.

\textbf{Blockwise Eikonal Supervision.}
TDM applies Eikonal supervision using a single geometry-derived speed target and the gradient of travel
time with respect to the complete configuration. Our rigid-body formulation instead applies the Eikonal
relation separately to the translational and rotational coordinate blocks using their corresponding geometric
targets. This blockwise regularizer supervises local motion difficulty in each component; it is not a derived
minimum-time equation for joint translation and rotation. Geometry supplies the target speeds $S_c^\star$,
while the gradient of the learned travel-time field gives the corresponding speed estimates $S_c$; the
blockwise Eikonal residual encourages consistency between these two quantities. The loss is applied at both
endpoints:
\begin{equation}
L_E^{\mathrm{shape}}=
\sum_{k,c}
w_E(S^\star_c(q_k))
\left(S^\star_c(q_k)\|\nabla_{q_k^c}T\|_2-1\right)^2.
\label{eq:block-eikonal}
\end{equation}
Here $k\in\{s,g\}$, $c\in\{\mathrm{trans},\mathrm{rot}\}$, $q_k^{\mathrm{trans}}=p_k$, $q_k^{\mathrm{rot}}=r_k$, and $\nabla_{q_k^c}T$ differentiates $T(q_s,q_g)$ with respect to that coordinate block. Lower target speeds require larger gradient magnitudes in the corresponding blocks of the same travel-time function; gradient norms are floored for numerical stability.

\textbf{Inverse-speed ($1/s$) Weighting.}
Differentiating a residual $(S^\star_c\|\nabla_{q^c}T\|_2-1)^2$ introduces a factor of $S^\star_c$, which can weaken the training signal at low target speeds. We compensate with $w_E(s)=\min(w_{\max},1/s)$ for $s<s_0$ and $w_E(s)=1$ otherwise, where $s$ is the target speed, $w_{\max}>0$ caps the weight, and $s_0$ defines the low-speed region.

\textbf{Relative Normal Loss.}
TDM introduces a direct normal-alignment loss that aligns the travel-time gradient with the gradient of its
geometry-derived speed field near obstacles. Our setting requires separate translational and rotational
geometric directions, and the rotational clearance surrogate does not provide an exact differentiable
clearance gradient. We therefore replace direct normal alignment with a relative speed constraint along the
component-specific normals $n_c$. Relative Normal Loss complements pointwise Eikonal supervision by
encouraging the network-implied speed to decrease when probing toward reduced clearance. We probe
each normal with the goal fixed. Let $S_c(q;q_g)=\|\nabla_{q^c}T(q,q_g)\|_2^{-1}$ be the network-implied speed, computed from the predicted travel-time gradient. Unlike the geometry-derived target $S_c^\star(q)$, it depends on the goal through $T$ and is not a separate network output. Its change under a probe of size $\varepsilon>0$ is
\begin{equation}
G_c=\frac{S_c(q_s+\varepsilon n_c;q_g)-S_c(q_s;q_g)}{\varepsilon}.
\label{eq:relative-speed}
\end{equation}
For target speeds below one, we penalize slopes above a negative threshold $\rho_c$:
\begin{equation}
L_{\mathrm{dir}}^c=\mathbf 1[S_c^\star(q_s)<1]
\left(\rho_c-\operatorname{clip}(G_c,\rho_c,G_{\max})\right)^2,
\label{eq:relative-normal}
\end{equation}
Here $\mathbf 1$ is an indicator and $G_{\max}$ is the upper slope cap, above which the penalty has zero direct slope gradient. This geometric regularizer remains active at the speed floor; it does not match derivatives of the clipped target.

\textbf{Training Objective.}
We retain TDM's causality weighting~\cite{ni2025physics}, also used in subsequent work~\cite{Ni2026Weakly-supervised}. The weight $L_C=\exp(-\lambda_C T)$, $\lambda_C>0$, emphasizes smaller predicted travel times:
\begin{equation}
L=\mathbb E_{(q_s,q_g)}\!\left[L_C\left(
\lambda_E L_E^{\mathrm{shape}}+\alpha\lambda_{\mathrm{dir}}\sum_c L_{\mathrm{dir}}^c\right)\right].
\label{eq:total-loss}
\end{equation}
Here $\lambda_E,\lambda_{\mathrm{dir}}$ weight the two losses, and $\alpha\in[0,1]$ limits the normal
term's relative contribution. Normal supervision is introduced gradually. Relative to TDM, we retain the
pairwise metric formulation and causality weighting, replace its single configuration-level Eikonal supervision
with our translation–rotation blockwise regularization, omit the temporal-difference loss, and replace direct
normal alignment with Relative Normal Loss.

\begin{table*}[!t]
\vspace*{8pt}
\centering
\begin{minipage}[t]{\dimexpr(\textwidth-\columnsep)/2\relax}
\centering
\caption{Learning ablations. Quantitative results cover 12 tasks; time and path metrics report mean $\pm$ standard deviation.}
\label{tab:ablation-results}
\scriptsize
\setlength{\tabcolsep}{2.5pt}
\resizebox{\linewidth}{!}{%
\renewcommand{\arraystretch}{1.12}
\begin{tabular}{@{}ccccc@{}}
\toprule
Variant & SR (\%) $\uparrow$ & Time (s) $\downarrow$ & Trans. $\downarrow$ & Rot. (rad) $\downarrow$ \\
\midrule
Full & 96.1 & $0.038 \pm 0.014$ & $0.763 \pm 0.303$ & $3.557 \pm 1.280$ \\
Single-speed & \multicolumn{4}{c}{Training did not converge} \\
Single-speed w/o RNL & 83.6 & $0.046 \pm 0.019$ & $0.802 \pm 0.314$ & $3.967 \pm 1.454$ \\
w/o Sep. Enc. & 85.5 & $0.023 \pm 0.010$ & $0.764 \pm 0.346$ & $3.733 \pm 1.553$ \\
w/o $1/s$ & 84.6 & $0.049 \pm 0.019$ & $0.945 \pm 0.404$ & $3.839 \pm 1.422$ \\
w/o RNL & 89.6 & $0.038 \pm 0.014$ & $0.696 \pm 0.274$ & $3.688 \pm 1.302$ \\
Full + MPPI & 84.6 & $0.022 \pm 0.008$ & $0.475 \pm 0.219$ & $2.518 \pm 1.044$ \\
\bottomrule
\end{tabular}}
\end{minipage}\hfill
\begin{minipage}[t]{\dimexpr(\textwidth-\columnsep)/2\relax}
\centering
\caption{Planner comparison on 993 non-linear queries with the largest start-goal distances. Failure causes may overlap.}
\label{tab:planner-ablation}
\scriptsize
\setlength{\tabcolsep}{3pt}
\renewcommand{\arraystretch}{1.12}
\begin{tabular}{@{}ccccc@{}}
\toprule
Field & Planner & SR (\%) $\uparrow$ & Coll. (\%) $\downarrow$ & Conv. (\%) $\downarrow$ \\
\midrule
Ours & Ours & 86.2 & 10.1 & 6.7 \\
Ours & MPPI & 62.1 & 33.5 & 15.9 \\
TDM & MPPI & 45.3 & 10.5 & 44.2 \\
TDM & Ours & 71.0 & 15.7 & 22.9 \\
\bottomrule
\end{tabular}
\end{minipage}
\end{table*}

\subsection{Sampling-based Planning with Speed-guided Steering}
We generate paths by repeatedly sampling, scoring, and updating short motions using the learned field. Paths grow from both endpoints, with candidate rollouts combined and speed-guided corrections applied at each iteration. Let $q$ be the endpoint being extended and $\bar q$ the opposite endpoint. We sample bounded, temporally correlated Gaussian increments to form rollouts $q_{i,h}$, where $i$ indexes candidates and $h=1,\ldots,H$ indexes steps, with $H>1$. Their first increments are $d_i=q_{i,1}-q$.

\textbf{Scoring Rollouts with Local Slowness.}
Scoring only the remaining travel time $T(q_{i,h},\bar q)$ favors candidates closer to the destination in the learned field, but does not account for the difficulty of reaching them from $q$. We therefore also evaluate $T(q,q_{i,h})$. Dividing this local travel-time estimate by displacement gives a slowness cost: a candidate with low speed is penalized even if its remaining cost is low. Define $\ell_{i,h}=\max(\|q_{i,h}-q\|_2,\epsilon_\ell)$ and $u_{i,h}=T(q,q_{i,h})/\ell_{i,h}$. The rollout score is
\begin{equation}
\begin{aligned}
F_{i,h}&=T(q_{i,h},\bar q)+\lambda_u u_{i,h},\\
J_i&=a_1F_{i,1}+\frac{a_H}{H-1}\sum_{h=2}^{H}F_{i,h},
\end{aligned}
\label{eq:local-planner-cost}
\end{equation}
where $\lambda_u>0$ balances local difficulty against remaining time. The weights $a_1,a_H>0$ balance the immediate step and the later rollout. Lower-scoring candidates receive larger exponential weights, combining their first increments as in sampling-based path-integral control~\cite{Williams2017InformationMPC}.

\textbf{Correcting the Step with Speed-guided Steering.}
The rollout score determines which candidates influence the update. Steering additionally uses how predicted speed varies across their directions to adjust the combined step. We reuse the first-step speed estimates $s_i=1/u_{i,1}$ and their mean $\bar s$, fitting speed deviations to pose increments:
\begin{equation}
\hat g=\arg\min_g\sum_i(s_i-\bar s-g^\top d_i)^2+\lambda_g\|g\|_2^2,
\label{eq:sampled-speed-fit}
\end{equation}
where $\lambda_g>0$ regularizes the fit. Although each $s_i$ is a scalar, comparing speeds across different $d_i$ estimates a direction of increasing speed in the joint translation and rotation coordinates. We normalize this direction as $\hat n=\hat g/\max(\|\hat g\|_2,\epsilon_g)$ and correct the weighted step:
\begin{equation}
\begin{aligned}
w_i&=\frac{\exp(-\beta_P J_i)}{\sum_j\exp(-\beta_P J_j)},\\
q^+&=q+\sum_i w_i d_i+b_P\delta\hat n,
\end{aligned}
\label{eq:planner}
\end{equation}
Here $\beta_P>0$ controls cost selectivity, and $b_P>0$ sets the correction strength. The sampled-increment bound $\delta>0$ also scales the correction; the final step may exceed this bound. The $\epsilon$ constants stabilize computation. Rotation updates use local increments in normalized rotation-vector coordinates.

We retain only this corrected first step and resample from $q^+$. Repeating the process alternately at both endpoints extends the paths until they meet. 

\begin{figure*}[t]
\vspace*{2pt}
\centering
\includegraphics[width=\textwidth]{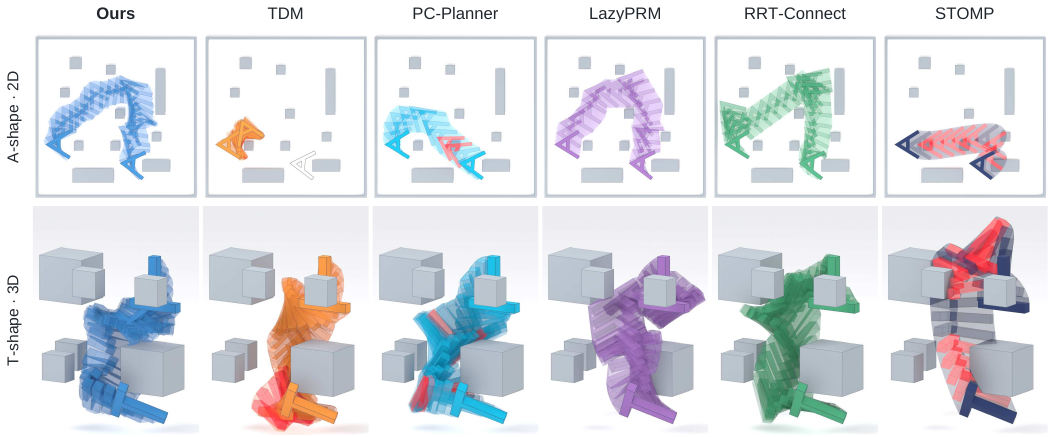}
\caption{Qualitative path comparison across six planners. Rows show A-shaped planar navigation and T-shaped spatial navigation. Columns show Ours, TDM, PC-Planner, LazyPRM, RRTConnect, and STOMP. Overlaid object poses show translation and rotation along each path; red poses indicate collisions. TDM fails to converge in the A-shaped tasks and collides in the T-shaped task. PC-Planner and STOMP produce collisions in both tasks.}
\label{fig:planner-paths}
\end{figure*}
\begin{table*}[t]
\centering
\begin{minipage}[t]{0.49\textwidth}
\centering
\caption{2D planning results (Overall).}
\label{tab:results-2d}
\scriptsize
\setlength{\tabcolsep}{2.5pt}
\renewcommand{\arraystretch}{1.12}
\begin{tabular}{@{}ccccc@{}}
\toprule
Method & SR (\%) $\uparrow$ & Time (s) $\downarrow$ & Trans. $\downarrow$ & Rot. (rad) $\downarrow$ \\
\midrule
Ours & $94.3$ & $0.048 \pm 0.026$ & $0.681 \pm 0.375$ & $2.837 \pm 1.574$ \\
TDM & $86.7$ & $0.097 \pm 0.052$ & $0.459 \pm 0.282$ & $2.355 \pm 1.490$ \\
RRTConnect & $98.9$ & $1.052 \pm 3.147$ & $0.780 \pm 0.506$ & $3.146 \pm 1.533$ \\
LazyPRM & $98.1$ & $1.670 \pm 3.375$ & $1.092 \pm 0.676$ & $2.751 \pm 1.674$ \\
PC-Planner & $84.2$ & $0.029 \pm 0.015$ & $0.449 \pm 0.262$ & $2.313 \pm 1.425$ \\
STOMP & $71.7$ & $0.071 \pm 0.259$ & $0.369 \pm 0.226$ & $2.168 \pm 1.483$ \\
\bottomrule
\end{tabular}
\end{minipage}
\hfill
\begin{minipage}[t]{0.49\textwidth}
\centering
\caption{3D planning results (Overall). }
\label{tab:results-3d}
\scriptsize
\setlength{\tabcolsep}{2.5pt}
\renewcommand{\arraystretch}{1.12}
\begin{tabular}{@{}ccccc@{}}
\toprule
Method & SR (\%) $\uparrow$ & Time (s) $\downarrow$ & Trans. $\downarrow$ & Rot. (rad) $\downarrow$ \\
\midrule
Ours & $92.9$ & $0.055 \pm 0.016$ & $1.117 \pm 0.328$ & $4.499 \pm 1.269$ \\
TDM & $63.7$ & $0.130 \pm 0.066$ & $0.631 \pm 0.251$ & $2.777 \pm 1.101$ \\
RRTConnect & $98.8$ & $2.241 \pm 3.628$ & $0.942 \pm 0.425$ & $4.521 \pm 1.937$ \\
LazyPRM & $72.4$ & $10.186 \pm 8.009$ & $2.557 \pm 1.387$ & $8.727 \pm 4.941$ \\
PC-Planner & $59.9$ & $0.046 \pm 0.016$ & $0.730 \pm 0.236$ & $2.867 \pm 1.363$ \\
STOMP & $70.5$ & $0.349 \pm 0.831$ & $0.671 \pm 0.203$ & $2.409 \pm 1.350$ \\
\bottomrule
\end{tabular}
\end{minipage}
\end{table*}

\section{Experiments}
\label{sec:experiments}

We evaluate Shape-NTFields through ablations, 2D and
3D rigid-body planning benchmarks, and closed-loop push-
ing with reactive replanning.

\subsection{Implementation details.}
All methods are evaluated on a workstation with an NVIDIA RTX 3090 GPU, an Intel Core i9-10980XE CPU at 3.00 GHz, and 128 GB RAM. For each shape and environment, Shape-NTFields is trained on 750,000 configuration pairs sampled independently of the test pairs, without expert trajectories. We use AdamW with weight decay of 0.5 and batch size of 2,000. The encoders and fusion network use hidden width 256. During planning, we sample 50 candidate rollouts with horizon $H=5$ and candidate-step bound $\delta=0.015$. Averaged across the 2D and 3D tasks, training takes approximately 1.98 h per model for Ours, 0.54 h for TDM, and 1.41 h for PC-Planner.

\subsection{Ablation Case Studies}
\label{sec:exp-ablation}
We isolate the contributions of the field representation, training objective, and planning strategy through two ablation studies: architecture and training variants with our planner fixed, and planner variants on the same learned fields. Success rate (SR) measures collision-free goal attainment. 

\textbf{Architecture and training.}
Across 12 tasks combining L, A, and V shapes with four 3D environments, we independently replace separate encoders with a single pose encoder (w/o Sep. Enc.), remove inverse-speed weighting (w/o $1/s$), or remove Relative Normal Loss (w/o RNL). With separate encoders retained, single-speed unified Eikonal training fails to converge with RNL but achieves 83.6\% SR without it. At the same RNL-disabled setting, blockwise dual-speed supervision achieves 89.6\%, a 6.0-point improvement. Table~\ref{tab:ablation-results} also evaluates the full field with TDM-style MPPI.

The encoder replacement and two loss removals reduce SR by 10.6, 11.5, and 6.5 percentage points, respectively. Relative Normal Loss improves success without increasing mean inference time. Length and rotation describe different successful subsets, so they do not isolate component effects.

\textbf{Planner.}
We compare our planner and TDM-style MPPI on both learned fields. After excluding collision-free straight-line interpolations, we retain the 25\% of remaining pairs with the largest start-goal distances, yielding 993 challenging queries. Collision and non-convergence may overlap in Table~\ref{tab:planner-ablation}, so their percentages need not sum to $100-\mathrm{SR}$.

With the field fixed, our planner improves SR over MPPI by 24.1 percentage points on our field and 25.7 points on TDM. On TDM, fewer queries fail to converge but more encounter collisions, indicating that goal progress alone does not ensure a safe path. Conversely, fixing our planner and replacing the TDM field with ours raises SR from 71.0\% to 86.2\% while reducing both failure rates. These results show complementary contributions from the learned field and planning strategy, with mean inference time increasing modestly from 0.022 to 0.038 s on our field.

\subsection{Baseline Comparison in 2D and 3D}
\label{sec:exp-benchmark}

\label{sec:exp-setup}

\textbf{Tasks and datasets.}
We evaluate seven shapes (rectangle, L, F, A, V, 4, and T) in four environments per dimension, using the same 200 valid start-goal pairs per task for all methods. 2D Planar tasks allow two translations and one rotation; 3D spatial tasks allow three translations and three rotations.

\textbf{Baselines.}
We compare with OMPL RRTConnect and LazyPRM~\cite{Kuffner2000RRT-connect:,Bohlin2000LazyPRM}, STOMP initialized by linear interpolation~\cite{Kalakrishnan2011STOMP:}, TDM with MPPI~\cite{ni2025physics}, and PC-Planner with bidirectional path extraction~\cite{Shen2024PC-Planner:}. LazyPRM retains its roadmap across queries for each fixed shape and environment. Every method has a 30-second budget; timeouts count as failures.

\textbf{Success criteria and metrics.}
Success requires reaching the goal within budget and passing ground-truth path collision checking. SR is averaged equally across tasks. Time, normalized translation length, and accumulated rotation report pooled means and standard deviations over successful queries. Translation is normalized by the longest environment side. Since path metrics use different successful subsets, they should be interpreted alongside SR. Neural planning times measure path generation, whereas search-based timings include internal collision checking. Tables~\ref{tab:results-2d} and~\ref{tab:results-3d} summarize the overall results.

\textbf{Planning performance.}
In 2D, Shape-NTFields achieves 94.3\% SR at 0.048 s mean inference time, above TDM (86.7\%), PC-Planner (84.2\%), and STOMP (71.7\%). RRTConnect and LazyPRM attain higher SRs of 98.9\% and 98.1\%, but require mean solve times of 1.052 and 1.670 s. PC-Planner is faster at 0.029 s but less successful at 84.2\%.

\textbf{2D to 3D.}
Shape-NTFields maintains similar performance as the pose space expands from planar to spatial planning: SR decreases only from 94.3\% to 92.9\%, while mean inference time increases from 0.048 to 0.055 s. RRTConnect and STOMP also retain similar SR, but their mean times more than double and quadruple, respectively. In contrast, TDM, PC-Planner, and LazyPRM lose 23.0, 24.3, and 25.7 percentage points in SR while also taking longer. Shape-NTFields therefore shows comparatively small degradation in both success and inference time.

\textbf{Qualitative paths.}
Figure~\ref{fig:planner-paths} shows representative A-shaped planar and V-shaped spatial trajectories. Shape-NTFields produces gradual pose changes, while several baselines exhibit collisions, or failure to reach the goal.
\begin{table}[t]
\vspace*{8pt}
\centering
\caption{Closed-loop planning and pushing control in simulation. Each environment includes 100 trials per shape (rectangle, L, A). Success requires goal attainment without collision.}
\label{tab:push-sim}
\small
\setlength{\tabcolsep}{8pt}
\renewcommand{\arraystretch}{1.12}
\begin{tabular}{@{}ccc@{}}
\toprule
Environment & SR (\%) $\uparrow$ & Collision (\%) $\downarrow$ \\
\midrule
1 & 78.0 & 22.0 \\
3 & 93.3 & 6.7 \\
4 & 89.3 & 10.7 \\
Overall & 86.9 & 13.1 \\
\bottomrule
\end{tabular}
\end{table}

\begin{figure}[t]
\centering
\includegraphics[width=\linewidth]{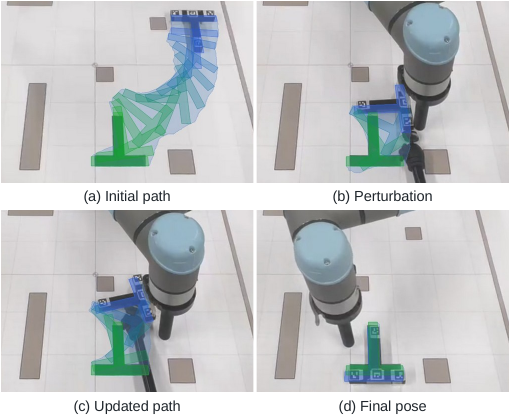}\vspace{-0.1in}
\caption{Real-world Push-T with an external perturbation. (a) Initial planned path. (b) Perturbation during execution. (c) Updated path from the measured block pose. (d) Final observed pose. The planner reuses the learned travel-time field within the repeated loop of perception, planning, and pushing.}
\label{fig:real-perturbation}\vspace{-.2in}
\end{figure}

\subsection{Closed-loop Pushing in Simulation}
\label{sec:exp-push-sim}

We next evaluate the learned field under imperfect non-prehensile pushing dynamics.

\textbf{Task and controller.}
We test rectangle, L, and A shapes in environments 1, 3, and 4, using 100 pairs per task (900 trials). A cylindrical pusher executes short actions, with pose measurement and replanning after each push. The controller selects an action toward a look-ahead path pose from 10,000 primitives motions calibrated per shape.

Obstacles exert no contact forces, so collisions are detected by per-frame object-footprint checks.

\textbf{Metrics and results.}
Of 900 trials, 899 reach the goal, while 86.9\% complete the task without collision and satisfy the success criterion (Table~\ref{tab:push-sim}). Collision-free SR ranges from 78.0\% to 93.3\% across environments. The gap between goal attainment and collision-free success indicates that most failures arise from execution-time collisions rather than lack of progress toward the goal. This experiment evaluates planning and control jointly, since pushing errors may cause collisions before replanning corrects the motion.

\subsection{Real-World Closed-Loop Pushing}
\label{sec:exp-real}

We evaluate the framework on a physical UR5e tabletop pushing task, using a RealSense D435 for block-pose estimation and the same controller as in simulation. Across eight real-world pushing trials across different start--goal configurations and achieve a 100\% success rate. The average planning time is 0.051s, comparable to the 0.048~s mean inference time in the planar simulation benchmark, showing that fast planning carries over to real-world deployment. The average task execution time is 121.7s. 

\subsection{Real-World Pushing under External Disturbances}
\label{sec:exp-perturbation}

We test recovery from unexpected object-pose changes by moving the T-block to a new collision-free pose during execution (Fig.~\ref{fig:real-perturbation}). The system replans from the displaced configuration and continues toward the original goal without retraining or restart. We conduct three disturbance trials with a 100\% success rate, achieving an average planning time of 0.051 s and task execution time of 215.4 s. These results demonstrate fast closed-loop replanning under both nominal and disturbed conditions.

\section{Conclusion}
\label{sec:conclusion}

We presented Shape-NTFields for rigid-body motion planning. Separate supervision of translation and rotation captures their different geometric restrictions, while a planner that considers local motion difficulty uses the learned field to guide progress toward the goal. Together, these designs yield high planning success with fast inference: on the 3D benchmark, Shape-NTFields achieves a 92.9\% success rate compared with 63.7\% for TDM. The ablations support contributions from both the learned field and the planning strategy. In UR5e pushing demonstrations, the same field supports repeated replanning from measured poses, allowing execution to continue after human perturbations. The current method learns a separate field for each fixed body and static environment; extending this reuse across shapes and changing obstacle geometries is a direction for future work.









\FloatBarrier
\bibliographystyle{IEEEtran}
\bibliography{references}

\end{document}